\documentclass[conference]{IEEEtran}
\IEEEoverridecommandlockouts
\usepackage{cite}
\usepackage{amsmath,amssymb,amsfonts}
\usepackage{algorithmic}
\usepackage{graphicx}
\usepackage{textcomp}
\usepackage{xcolor}
\usepackage{url}
\usepackage{multirow}
\def\BibTeX{{\rm B\kern-.05em{\sc i\kern-.025em b}\kern-.08em
    T\kern-.1667em\lower.7ex\hbox{E}\kern-.125emX}}
\begin{document}

\title{Distill2Vec: Dynamic Graph Representation Learning with Knowledge Distillation}

\author{\IEEEauthorblockN{Stefanos Antaris}
\IEEEauthorblockA{KTH Royal Institute of Technology \\
Hive Streaming AB \\
Sweden \\
antaris@kth.se}
\and
\IEEEauthorblockN{Dimitrios Rafailidis}
\IEEEauthorblockA{Maastricht University \\
Netherlands\\
dimitrios.rafailidis@maastrichtuniversity.nl}
}

\maketitle

\begin{abstract}
Dynamic graph representation learning strategies are based on different neural architectures to capture the graph evolution over time. However, the underlying neural architectures require a large amount of parameters to train and suffer from high online inference latency, that is several model parameters have to be updated when new data arrive online. In this study we propose Distill2Vec, a knowledge distillation strategy to train a compact model with a low number of trainable parameters, so as to reduce the latency of online inference and maintain the model accuracy high. We design a distillation loss function based on Kullback-Leibler divergence to transfer the acquired knowledge from a teacher model trained on offline data, to a small-size student model for online data. Our experiments with publicly available datasets show the superiority of our proposed model over several state-of-the-art approaches with relative gains up to 5\% in the link prediction task. In addition, we demonstrate the effectiveness of our knowledge distillation strategy, in terms of number of required parameters, where Distill2Vec achieves a compression ratio up to 7:100 when compared with baseline approaches. For reproduction purposes, our implementation is publicly available at \url{https://stefanosantaris.github.io/Distill2Vec}.
\end{abstract}


\section{Introduction}

Graph representation learning is a fundamental problem, with ubiquitous applicability in various real-world domains, such as social networks \cite{Zhu2017, velickovic2018}, biological protein-protein networks \cite{hamilton2017, Fout2017}, recommender systems \cite{Cao2019, Goyal2018ht}, and so on. The main objective is to learn low-dimensional dense vector representations - node embeddings that capture the structural and content information of each node \cite{hamilton17, Liu2018}. Neighbor nodes or nodes with similar interests e.g., similar movie preferences, are mapped to vectors with close proximity in a latent embedding space. The learned node embeddings have been proven beneficial for a wide variety of machine learning tasks such as predictions of future friendships between users in  social networks \cite{Geng2015}, and recommendations of new products to customers in E-commerce platforms \cite{Zhao2019}. 

Early graph representation learning approaches mainly focus on static graphs \cite{Grover2016, kipf2017, Perozzi2014, velickovic2018}. However, most real-world applications are dynamic. Static approaches completely ignore the temporal aspect of the graph. To efficiently capture the evolution in the latent embedding space, dynamic graph representation learning approaches compute node embeddings based on a sequence of graph snapshots at different time steps \cite{Sankar2020, Zhou2018, Zhu2017, Trivedi2019DyRepLR}. Existing approaches explore several techniques to accurately learn node embeddings, such as temporal regularizers \cite{mahdavi2019, Zhou2018}, Recurrent Neural Networks \cite{Trivedi2019DyRepLR, Hajiramezanali2019}, and joint-self attention mechanisms \cite{Sankar2020}.

Although dynamic graph representation learning strategies produce accurate predictions, they are based on deep neural network architectures with a large number of model parameters. Moreover, the number of parameters significantly increases by several orders of magnitude, along with the number of graph snapshots. Due to the vast amount of model parameters such approaches incur high online inference latency, which prohibits their direct applications into a real-world setting with almost real-time response requirements~\cite{Bucilua2006, hinton2015, Liu_2019_CVPR, Mary2019, Jiaxi2018}. For example, the model size negatively impacts the performance of  recommendation systems in social networks, where predictions have to be calculated in real time~\cite{Jiaxi2018,Li2017}.

Knowledge distillation is a model independent strategy to generate compact models that exhibit low online inference latency. \cite{Bucilua2006, hinton2015}. The basic idea of knowledge distillation is to train a large model, namely \textit{teacher}, as an offline process. The teacher model can employ computationally expensive deep neural networks, as there are no strict requirements on latency and computational resources during offline learning. Having trained the teacher model, the knowledge can be transferred to a smaller model, namely \textit{student}, by reducing the model size. Therefore, the student model can be deployed to online applications, satisfying the low online inference latency requirements \cite{Jiaxi2018, Mary2019, Cao2019}. However, the impact of knowledge distillation on graph representation learning for dynamic graphs has not been studied so far.

In this paper, we propose a knowledge distillation strategy, namely Distill2Vec, to generate a compact student model with low online inference latency for graph representation learning on dynamic graphs. The teacher model learns the latent node representations by employing a self-attention mechanism on the offline graph snapshots.  To train a smaller student model on the online graph snapshots, we formulate a distillation loss function, allowing the student model to distill the knowledge of the pretrained teacher model. In doing so, the student model can generate similar predictions as the teacher model, while significantly reducing the model parameters. Our main contributions are summarized as follows:
\begin{itemize}
    \item We propose Distill2Vec, a knowledge distillation strategy on dynamic graph representation learning approaches. We formulate a distillation loss function based on Kullback-Leibler divergence to transfer the knowledge from the teacher model on the offline data, to a smaller student model when learning online data. In addition, Distill2Vec employs a self-attention mechanism to capture the graph evolution in the learned node embeddings.
    \item We demonstrate that the student model significantly reduces the online inference latency, in terms of the number of trainable parameters, when compared with the teacher model. Moreover, the proposed student model overcomes any bias introduced by the pretrained teacher model, achieving high accuracy in the online link prediction task.
\end{itemize}
Our experiments on two real-world dynamic networks demonstrate the superiority of our proposed knowledge distillation strategy, against several state-of-the-art methods. 

The remainder of the paper is organized as follows: Section \ref{sec:rel_work} reviews the related work and in Section \ref{sec:model} we describe the proposed knowledge distillation strategy. The experimental evaluation is presented in Section \ref{sec:exp} and we conclude the study in Section \ref{sec:conclusion}.

\section{Related Work} \label{sec:rel_work}

Static graph representation learning approaches exploit matrix factorization techniques \cite{Cao2015, Ou2016} and random walks \cite{Perozzi2014, Grover2016} to learn accurate node embeddings. With the advent of deep learning methods, several graph neural network approaches have been proposed, such as Graph Convolutional Networks (GCN) \cite{kipf2016variational, hamilton2017}, and Graph Attentions \cite{velickovic2018}. However, these methods are designed to learn node embeddings for static graphs and do not reflect on the dynamic setting of real-world applications.

Dynamic graph representation learning approaches aim to compute accurate node embeddings capturing the graph evolution. For example, early attempts on dynamic graph representation learning employ temporal smoothness techniques to calculate similar latent representations between consecutive graph snapshots \cite{Sarkar2005, Zhu2017}. DynamicTriad exploits the triadic closure process as a smoothness guidance to identify the temporal patterns of social networks \cite{Zhou2018}. DyREP models the occurrence of an edge as a point process and captures the interleaved dynamics between consecutive graph snapshots \cite{Trivedi2019DyRepLR}. DynGem employs auto-encoders to compute the latent node embeddings for each graph snapshot \cite{goyal2018dyngem}. To ensure smoothness, the node embeddings of each autoencoder are initialized based on the pretrained node embeddings of the previous graph snapshot. Similarly, Dynamic Joint Variational Graph AutoEncoder (DynVGAE) shares parameter weights between consecutive variational graph auto-encoders \cite{kipf2016variational} \cite{mahdavi2019}. Recently, dynamic graph representation learning approaches summarize the historical graph snapshots in the hidden states of recurrent neural networks \cite{Pareja2020, Goyal2020}. These approaches scale poorly along with the increase of the number of graph snapshots. To incorporate the network temporal information in the Graph Neural Networks (GNN), Temporal Dependent Graph Neural Network (TDGNN) employs aggregation functions on the neighbor nodes between consecutive graph snapshots. Dynamic Self-Attention (DySAT) applies a self-attention mechanism to encode the structural and temporal dynamics of each node over consecutive graph snapshots \cite{Sankar2020}. However, existing state-of-the-art approaches are based on neural architectures with large model sizes and cannot capture the evolution of the graph without encountering prohibitive online training costs, as we will show later in Section~\ref{sec:exp}. 

A recent attempt to reduce the model sizes of graph representation learning approaches based on the DMTKG knowledge distillation strategy, presented in \cite{Jiaqi2019}. DMTKG employs Heat Kernel Signatures (HKS) to extract the nodes' descriptors and thereafter forward the nodes' descriptors to GCNs to learn the latent node representations. The knowledge of the large teacher model is distilled to the compact student model through a distillation loss function based on the weighted cross entropy. However, DMTKG is designed to learn node embeddings on static graphs, ignoring the temporal evolution of dynamic graphs.

\section{Proposed Model} \label{sec:model}

We define a dynamic graph as a sequence of $T$ graph snapshots $\mathcal{G} = \{\mathcal{G}_1, \ldots,\mathcal{G}_T\}$. For each time step $t=1,\ldots,T$, the graph snapshot $\mathcal{G}_t = (\mathcal{V}_t, \mathcal{E}_t)$ is an undirected graph, where $\mathcal{V}_t$ is the set of $n_t=|\mathcal{V}_t|$ nodes and $\mathcal{E}_t$ corresponds to the set of links. The goal of dynamic graph representation learning is to map each node $u \in \mathcal{V}_t$ to $d$-dimensional node embeddings  $\mathbf{H}_t (u) \in \mathbb{R}^{d}$, with $d \ll n_t$, at the time step $t$~\cite{Goyal2018, hamilton17, Grover2016}. Provided that graphs evolve over time, the node embedding $\mathbf{H}_t(u)$ should encode the evolution of the node $u \in \mathcal{V}_t$ over a specific window size $l$ of consecutive graph snapshots $\{\mathcal{G}_{t-l},\ldots, \mathcal{G}_t\}$, up until the $t$-th time step. Note that in our model we do not consider all the previous time steps, as large values of the window size $l$ introduce noise to the node embeddings,  degrading the performance of the graph representation learning approaches as we will demonstrate in Section \ref{sec:param_sens}.

Dynamic graph representation learning models employ deep neural network architectures to learn accurate node embeddings, at the cost of high online inference latency \cite{Goyal2018, mahdavi2019, Sankar2020}. The goal of our knowledge distillation strategy is to generate a compact student model $\mathcal{S}$ with low online inference latency, and retain the accuracy of the pretrained large teacher model $\mathcal{T}$ \cite{anil2018, Jimmy2014, hinton2015, Bucilua2006}. In particular, the teacher model $\mathcal{T}$ is pretrained to learn the node embeddings $\mathbf{H}^{\mathcal{T}}$ on the offline graph snapshots. We denote by $\mathcal{G}^{\mathcal{T}} = \{\mathcal{G}_1, \ldots, \mathcal{G}_m\}$ the $m$ consecutive offline graph snapshots of the dynamic graph $\mathcal{G}$, with $1 \leq m < T$. Thereafter, the student model $\mathcal{S}$ exploits the teacher model $\mathcal{T}$, to learn accurate node representations $\mathbf{H}^{\mathcal{S}}$ on the online graph snapshots $\mathcal{G}^{\mathcal{S}} = \{\mathcal{G}_{m+1}, \ldots, \mathcal{G}_{T}\}$. To transfer the knowledge of the pretrained teacher model $\mathcal{T}$, the student model $\mathcal{S}$ minimizes a distillation loss function $L^{\mathcal{D}}$ \cite{Bucilua2006, hinton2015}. The distillation loss function $L^{\mathcal{D}}$ calculates the prediction error of the student model $\mathcal{S}$ on the online graph snapshots, and the deviation of $\mathbf{H}^{\mathcal{S}}$ from the node embeddings $\mathbf{H}^{\mathcal{T}}$. In Section \ref{sec:teacher}, we describe the offline teacher model Distill2Vec-$\mathcal{T}$, and then in Section \ref{sec:student} we present the knowledge distillation strategy of the online student model Distill2Vec-$\mathcal{S}$.

\subsection{Distill2Vec-$\mathcal{T}$ - Teacher Model} \label{sec:teacher}

The teacher model Distill2Vec-$\mathcal{T}$ learns the latent representations $\mathbf{H}^{\mathcal{T}}_t$ based on the $m$ offline graph snapshots $\mathcal{G}^{\mathcal{T}}$. Distill2Vec-$\mathcal{T}$ employs two self-attention layers~\cite{velickovic2018,Sankar2020,vaswani2017attention}. The first layer, namely structural self-attention, captures the structural properties of each node $u \in \mathcal{V}_t$ at the $t$-th graph snapshot. The second layer, namely temporal self-attention, models the evolution of the graph, given a sequence of $l$ graph snapshots $\{\mathcal{G}_{t-l}, \ldots, \mathcal{G}_t\}$. Provided that the teacher model Distill2Vec-$\mathcal{T}$ is trained as an offline process, we consider all the graph snapshots $\mathcal{G}^{\mathcal{T}}$ for the temporal self-attention layer ($l=m$). The input of the structural self-attention layer at the $t$-th time step is the set of input node representations $\mathbf{X}_t \in \mathbb{R}^{n_t \times n_t}$, where $\mathbf{X}_t(u)$ is the one-hot encoded vector of the node $u \in \mathcal{V}_t$. The output is a $d$-dimensional structural node representation $\mathbf{Z}_t(u) \in \mathbb{R}^d$, calculated as follows:

\begin{equation}
\mathbf{Z}_t(u) = ELU\bigg( \displaystyle\sum_{u \in \mathcal{N}_t(u)} \alpha_t(u,v) \mathbf{W}_t \mathbf{X}_t(u)\bigg)
\end{equation}
where $\mathcal{N}_t(u)$ is the neighborhood set of the node $u$ at the $t$-th time step, $\mathbf{W}_t \in \mathbb{R}^{d \times n_t}$ is the weight transformation matrix for each input node representation $\mathbf{X}_t(u)$, and ELU is the exponential linear unit activation function. Variable $\alpha_t(u,v)$ corresponds to the learned coefficients, calculated based on the softmax over the neighbors of each node $u$, as follows:

\begin{equation}
\alpha_t(u,v) = \frac{exp\big(e_t(u,v)\big)}{\displaystyle\sum_{w \in \mathcal{N}_t(u)}exp\big(e_t(u,w)\big)} 
\end{equation}
$$\text{with} \quad e_t(u,v) = f\big( A_t(u,v) \cdot \mathbf{a}_t^{\top} [\mathbf{W}_t \mathbf{X}_t(u) \parallel \mathbf{W}_t \mathbf{X}_t(v)]\big)
$$
$f$ is the LeakyRelu activation function, $\mathbf{a}_t \in \mathbb{R}^{2n_t}$ is a $2n_t$-dimensional weight vector parameterizing the attention process between nodes $u$ and $v$, and $\parallel$ denotes the concatenation operation. The attention weight $e_t(u,v)$ indicates the contribution of the node $v$ to the node $u$ at the $t$-th time step \cite{Sankar2020, velickovic2018}.

Having computed the $d$-dimensional structural node representations $\mathbf{Z}_t$ for each time step $t=1,\ldots,m$, we capture the graph evolution in the temporal attention layer. In contrast to the structural attention layer that learns the structural properties of the nodes at each time step, the temporal attention layer emphasizes on the evolution of each node over $l$ consecutive graph snapshots, with $l=m$ for the teacher model. The input of the temporal attention layer, denoted by $\mathbf{X'}_t(u) \in \mathbb{R}^{l \times d}$, is calculated as follows $\mathbf{X'}_t(u) = Concat(\mathbf{Z}_{t-l}(u), \ldots, \mathbf{Z}_t(u))$, that is the concatenation of the $l$ structural node representations of each node $u$. We apply the scaled dot-product form of attention \cite{Sankar2020,vaswani2017attention}, where the structural node representations $\mathbf{X'}_t$ are the queries, keys and values of the attention process. For each node $u \in \mathcal{V}_t$, the temporal attention layer calculates $l$ new $k$-dimensional representations $\mathbf{B}_t(u) \in \mathbb{R}^{l \times k}$ as follows: 

\begin{equation}
    \mathbf{B}_t(u) = \boldsymbol{\beta}_t(u)(\mathbf{X'}_t(u)\mathbf{W}_t^{value})
\end{equation}
$\mathbf{W}_t^{value} \in \mathbb{R}^{d \times k}$ is the linear projection matrix of the structural node representations of each node $u$. Variable $\boldsymbol{\beta}_t(u) \in \mathbb{R}^{l \times l}$ is the attention weight matrix that indicates the similarity of the node's $u$ structural embeddings in different graph snapshots. For each graph snapshot $i = t-l, \ldots,t$ and $j = t-l, \ldots, t$, we calculate the attention weight of the node $u$ as follows:

\begin{equation} 
\label{eq:cof}
     \beta^{ij}(u) = \frac{exp(c^{ij}(u))}{\displaystyle\sum^{t}_{r=t-l} exp(c^{ir})(u)} 
\end{equation}
$$     \text{with} \quad c^{ij}(u) = \bigg( \frac{((\mathbf{X'}_i(u) \mathbf{W}^{query})(\mathbf{X'}_j(u) \mathbf{W}^{key}))^{ij}}{\sqrt{k}} + M^{ij} \bigg)$$
$\mathbf{W}^{query} \in \mathbb{R}^{d \times k}$ and $\mathbf{W}^{key} \in \mathbb{R}^{d \times k}$ are the weight parameter matrices to transform the query and key input node representations, respectively \cite{vaswani2017attention}. A high attention weight $\beta^{ij}(u)$ corresponds to similar structural node embeddings for the node $u$ in the graph snapshots $\mathcal{G}^{\mathcal{T}}_i$ and $\mathcal{G}^{\mathcal{T}}_j$. In Equation \ref{eq:cof} $\mathbf{M} \in \mathbb{R}^{l \times l}$ is a mask matrix to encode the temporal order between different time steps $i$ and $j$. The values of the matrix $\mathbf{M}$ are defined as follows:

\begin{equation}
M^{ij} =
  \begin{cases}
    0       & \quad \text{if } i < j\\
    -\infty  & \quad \text{otherwise} 
  \end{cases}
\end{equation}

We employ multi-head attention on both the structural and temporal attention layers, to capture the evolution of different latent facets over time for each node $u \in \mathcal{V}_t$~\cite{Sankar2020}. The output of the multi-head attention on the structural attention layer is computed as follows:

\begin{equation} 
\label{eq:structural}
    \mathbf{C}_t(u) = Concat(\mathbf{Z}^1_t(u), \ldots, \mathbf{Z}^h_t(u))
\end{equation}
where $h$ is the number of attention heads and $\mathbf{C}_t(u) \in \mathbb{R}^d$ is the output representation of the node $u$ at the $t$-th time step. Similar to the structural attention layer, the output of the multi-head attention on the temporal attention layer is defined as follows:
\begin{equation}
\label{eq:temporal}
    \mathbf{D}_t(u) = Concat(\mathbf{B}^1(u), \ldots, \mathbf{B}^g_t(u))
\end{equation}
where $g$ is the number of attention heads applied to the temporal attention layer and $\mathbf{B}_t(u) \in \mathbb{R}^{l \times k}$ is the output node representations of the node $u$.

Having computed both the structural and the temporal node representations, we can calculate the final node representation $\mathbf{H}_t(u)$ for each node $u \in \mathcal{V}_t$. We encode the ordering information in the node representations $\mathbf{D}_t(u)$ of the temporal attention layer, by calculating the position embeddings $\mathbf{P}_t(u) \in \mathbb{R}^{d}$ for each node $u$ \cite{Gehring2017}. The final node representations $\mathbf{H}_t^{\mathcal{T}}(u)$ of the teacher model Distill2Vec-$\mathcal{T}$ are then computed by combining the output node representations $\mathbf{C}_t(u)$ of the structural attention layer with the position embeddings $\mathbf{P}_t(u)$ as follows:
\begin{equation}
\label{eq:embed}
 \mathbf{H}_t^{\mathcal{T}}(u) = \mathbf{C}_t(u) + \mathbf{P}_t(u)
\end{equation}
 
To train the teacher model and learn the node embeddings, we adopt the binary cross-entropy loss function with respect to the node embeddings $\mathbf{H}_t^{\mathcal{T}}(u)$:

\begin{equation}
\label{eq:loss}
\begin{array}{c}
    \displaystyle \min_{\mathbf{H}^{\mathcal{T}}_t} L = \displaystyle\sum_{u \in \mathcal{V}_t} \bigg( \displaystyle\sum_{v \in \mathcal{N}^{\text{walk}}_{t}(u)} -log \big(\sigma (<\mathbf{H}_t^{\mathcal{T}}(u), \mathbf{H}_t^{\mathcal{T}}(v)>)\big) \\
    -w_{neg} \cdot \displaystyle\sum_{u' \in \mathcal{P}^t_{neg}(u)} log \big( 1 - \sigma (<\mathbf{H}_t^{\mathcal{T}}(u'), \mathbf{H}_t^{\mathcal{T}}(u)>)\big)\bigg) 
\end{array}
\end{equation}
where $\sigma$ is the sigmoid activation function, $<,>$ is the inner product operation between node representations $\mathbf{H}^{\mathcal{T}}_t(u)$ and $\mathbf{H}^{\mathcal{T}}_t(v)$. $\mathcal{N}^{\text{walk}}_{t}(u)$ is the set of nodes explored in a fixed length random-walk started at the node $u$ at the $t$-th graph snapshot $\mathcal{G}_t$. $\mathcal{P}^t_{neg}(u)$ is a negative sampling distribution for the graph snapshot $\mathcal{G}_t$, and $w_{neg}$ is the negative sampling ratio. We optimize the weight parameter matrices in the structural and the temporal attention layers based on the loss function in Equation \ref{eq:loss} and the backpropagation algorithm.

\subsection{Distill2Vec-$\mathcal{S}$ - Student Model} \label{sec:student}

To reduce the high online inference latency of the teacher model Distill2Vec-$\mathcal{T}$, we train a compact student model Distill2Vec-$\mathcal{S}$ on the online graph snapshots $\mathcal{G}^{\mathcal{S}}$. For each time step $t = m+1, \ldots, T$, the student model Distill2Vec-$\mathcal{S}$ computes the structural node representations $\mathbf{C}_t(u)$, based on Equation \ref{eq:structural}. To capture the graph evolution over the last $l$ consecutive historical graph snapshots $\{\mathcal{G}^{\mathcal{S}}_{t-l}, \ldots, \mathcal{G}^{\mathcal{S}}_t\}$, Distill2Vec-$\mathcal{S}$ computes the temporal node representations $\mathbf{D}_t(u)$ according to Equation \ref{eq:temporal}. The final node representations $\mathbf{H}_t^{\mathcal{S}}(u)$ are calculated based on Equation \ref{eq:embed}.

We employ a knowledge distillation strategy on the student model Distill2Vec-$\mathcal{S}$ to transfer the knowledge of the pretrained teacher model Distill2Vec-$\mathcal{T}$. In practice, the student model Distill2Vec-$\mathcal{S}$ adopts the following distillation loss function $L^\mathcal{D}$ during the online training process:

\begin{equation}
    \label{eq:distil}
    \displaystyle \min_{\mathbf{H}^{\mathcal{S}}}L^\mathcal{D} = (1 - \gamma) L^\mathcal{S} + \gamma L^\mathcal{F}
\end{equation}
where $L^\mathcal{S}$ is the binary cross-entropy loss that measures the accuracy error of the student model on the online data, and $L^\mathcal{F} = \mathcal{KL}(H_t^{\mathcal{S}}(u) \mid H_t^{\mathcal{T}}(u))$ is the Kullback-Leibler (KL) divergence between the node embeddings $\mathbf{H}_t^{\mathcal{S}}(u)$ and $\mathbf{H}_t^{\mathcal{T}}(u)$ for each node $u \in \mathcal{V}_t$ \cite{Tian2020Contrastive}. This means that the student model Distill2Vec-$\mathcal{S}$ mimics the teacher model Distill2Vec-$\mathcal{T}$ during online training, to achieve similar performance with low number of model parameters \cite{Bucilua2006, Mary2019, Jiaxi2018}. Hyperparameter $\gamma \in [0,1]$ balances the distillation process and the prediction error of the student model Distill2Vec-$\mathcal{S}$ on the online data. High values of $\gamma$ reflect on generating node embeddings $\mathbf{H}^{\mathcal{S}}_t(u)$ similar to the node embeddings $\mathbf{H}^{\mathcal{T}}_t(u)$ of the student model Distill2Vec-$\mathcal{T}$. Instead, low values of $\gamma$ emphasize on the prediction errors of the student model Distill2Vec-$\mathcal{S}$. This allows the student model to overcome any bias introduced by the teacher and achieve similar or better performance than Distill2Vec-$\mathcal{T}$ \cite{Bucilua2006, hinton2015, Mary2019, Jiaxi2018}.

\section{Experiments} \label{sec:exp}

\subsection{Datasets}

We evaluate the performance of the proposed distillation strategy based on two publicly available datasets, that is the Yelp\footnote{\url{https://www.yelp.com/dataset}} and ML-10M\footnote{\url{https://grouplens.org/datasets/movielens/}} datasets. 
\begin{itemize}
    \item The Yelp dataset is a bipartite network with $6,569$ users and businesses and $95,361$ ratings. It consists of $16$ graph snapshots, where each graph snapshot contains the users/businesses and ratings within a $6$ month period.
    \item In ML-10M, the dynamic graph consists of $12$ graph snapshots with $20,537$ users/movies and $43,760$ user/tag interactions in MovieLens. Each graph snapshot corresponds to the user/tag interactions occurred within a $3$ month period.
\end{itemize}

\subsection{Evaluation Protocol}

We evaluate the performance of our proposed knowledge distillation strategy on the link prediction task. In our experiments, we train the teacher model Distill2Vec-$\mathcal{T}$ on the offline graph snapshots $\mathcal{G}^{\mathcal{T}}$. For each dataset, we consider the first $5$ time steps ($m=5$) as the offline graph snapshots $\mathcal{G}^{\mathcal{T}}$ and the remaining time steps as the online graph snapshots $\mathcal{G}^{\mathcal{S}}$, that is 11 and 7 test graph snapshots for the Yelp and ML-10M datasets, respectively. The student model Distill2Vec-$\mathcal{S}$ learns the node embeddings $\mathbf{H}^{\mathcal{S}}_t$ at the $t$-th time step based on $l$ consecutive online graph snapshots $\{\mathcal{G}^{\mathcal{S}}_{t},\ldots,\mathcal{G}^{\mathcal{S}}_{t-l} \}$. 

The task is to predict the unobserved links $\mathcal{O}_{t+1} = \mathcal{E}_{t+1}\setminus \{\mathcal{E}_{t}, \ldots, \mathcal{E}_{t-l}\}$ of the graph snapshot $\mathcal{G}^{\mathcal{S}}_{t+1}$,. Following the evaluation protocol of \cite{Sankar2020,mahdavi2019, Grover2016}, we compute a feature vector for a pair nodes $u \in \mathcal{V}_t$ and $v \in \mathcal{V}_t$ based on the node embeddings $\mathbf{H}^{\mathcal{S}}_t(u)$ and $\mathbf{H}^{\mathcal{S}}_t(v)$, and the Hadamard operator. We train a logistic regression classifier with evaluation links $o_{t+1}(u,v) \in \mathcal{O}_{t+1}$ for each node $u \in \mathcal{V}_{t+1}$ and $v\in\mathcal{V}_{t+1}$ from the graph snapshot $\mathcal{G}^{\mathcal{S}}_{t+1}$ and an equal number of randomly selected non-existing links $o_{t+1}(u,v) \notin \mathcal{O}_{t+1}$ for negative sampling \cite{Grover2016,Zhou2018, Sankar2020}. We hold the $20\%$ of the evaluation links $o_{t+1}(u,v) \in \mathcal{O}_{t+1}$ for validation set to tune the hyper-parameters of each model. From the remaining links, we randomly sample $60\%$ for training and keep the rest as a test set to evaluate the performance of the models.

We measure the online inference efficiency based on the required number of parameters to train each model. We adopt the Area Under the ROC Curve (AUC), to evaluate the performance of the link prediction task \cite{Sankar2020, Grover2016}. For each graph snapshot in $\mathcal{G}^{\mathcal{S}}$, we report average AUC values over five randomized runs.

\subsection{Baselines}

We compare the proposed Distill2Vec-$\mathcal{T}$ and Distill2Vec-$\mathcal{S}$ models with the following baseline strategies:

\begin{itemize}
    \item DynVGAE \cite{mahdavi2019}: a dynamic graph representation learning approach that exploits $l$ consecutive graph autoencoders~\cite{kipf2016variational} with shared trainable parameters. As there is no publicly available implementation, we published our source code of the DynVGAE model\footnote{\url{https://github.com/stefanosantaris/DynVGAE}}.
    \item DynamicTriad\footnote{\url{https://github.com/luckiezhou/DynamicTriad}} \cite{Zhou2018}: a deep neural network approach that employs triadic closure to capture the structural properties of the graph and temporal smoothness.
    \item TDGNN \cite{Qu2020}: a graph neural network extension that incorporates the nodes' structural information and edges' temporal evolution via edge aggregation functions. We implemented TDGNN from scratch and made our source code publicly available\footnote{\url{https://gith ub.com/stefanosantaris/TDGNN}}.
    \item DyREP\footnote{\url{https://github.com/uoguelph-mlrg/LDG}} \cite{Trivedi2019DyRepLR}: a two-time scale process that captures the temporal node interactions by employing deep recurrent model, so as to calculate the probability of occurrence of future links between two nodes.
    \item DMTKG-$\mathcal{T}$ \cite{Jiaqi2019}: the teacher model of the knowledge distillation strategy applied on the DeepGraph graph representation learning approach \cite{li2016deepgraph}. DMTKG-$\mathcal{T}$ computes the node embeddings on static graphs by employing Convolutional Neural Networks on the intermediate node representations generated by the Heat Kernel Signature (HKS). As the source code of DMTKG-$\mathcal{T}$ is not publicly available, we provide our implementation\footnote{\url{https://github.com/stefanosantaris/DMTKG}} for reproduction purposes.
    \item DMTKG-$\mathcal{S}$ \cite{Jiaqi2019}: the student model of the DMTKG knowledge distillation strategy. The student model distills the knowledge of the pretrained teacher model by employing a distillation loss function based on the weighted cross entropy.
    \item Distill2Vec-$\mathcal{L}$: a variant of the proposed student model, where we replace the Kullback-Leibler divergence $L^\mathcal{F}$ in Equation \ref{eq:distil} with the binary cross-entropy loss function, as in \cite{egad2020}. 
\end{itemize}

\subsection{Parameter Settings}
We tuned the hyper-parameters of each examined model following a cross-validation strategy. In Table \ref{tab:hyperparam}, we present the concluded values of the hyper-parameters of each model. In our experiments, we optimized the weight parameters of each model, employing the mini-batch gradient descent with the Adam optimizer \cite{kingma2014}. We initialized the learning rate to $1e-03$ and train each model for $200$ epochs. All experiments were performed on an Intel(R) Xeon(R) Bronze 3106 CPU 1.70GHz machine and GPU accelerated with the GEFORCE RTX 2080 Ti graph card.  

\begin{table}[h]
    \centering
    \caption{Concluded hyper-parameters of each examined model}
    \vspace{-0.2cm}
    \resizebox{.48\textwidth}{!}{
    \begin{tabular}{|c|c|c|c|c|c|c|}
    \hline
                        & \multicolumn{3}{c|}{\textbf{Yelp}} & \multicolumn{3}{c|}{\textbf{ML-10M}} \\ \hline \hline
        \multirow{2}{*}{\textbf{Model}}  & \textbf{Embedding} & \textbf{Window} & \textbf{Heads} & \textbf{Embedding} & \textbf{Window} & \textbf{Heads} \\
        & $d$ & $l$ & $h$/$c$ & $d$ & $l$ & $h$/$g$ \\ \hline
        \textbf{Distill2Vec}-$\mathcal{T}$ & 256 & 5 & 16 & 512 & 5 & 8/8 \\ \hline
        \textbf{Distill2Vec}-$\mathcal{S}$ & 64 & 2 & 2 & 128 & 2 & 4/4 \\ \hline
        \textbf{Distill2Vec}-$\mathcal{L}$ & 64 & 2 & 2 & 128 & 2 & 4/4 \\ \hline
        \textbf{DynVGAE} & 256 & 3 & N/A & 128 & 2 & N/A \\ \hline
        \textbf{DynamicTriad} & 256 & 3 & N/A & 512 & 2 & N/A \\ \hline
        \textbf{TDGNN} & 512 & 2 & N/A & 256 & 3 & N/A\\ \hline
        \textbf{DyREP} & 128 & 3 & N/A & 256 & 3 & N/A \\ \hline
        \textbf{DMTKG-$\mathcal{T}$} & 512 & N/A & N/A & 256 & N/A & N/A \\ \hline
        \textbf{DMTKG-$\mathcal{S}$} & 256 & N/A & N/A & 64 & N/A & N/A \\ \hline
    \end{tabular}
    }
    
    \label{tab:hyperparam}
\end{table}

\subsection{Performance Evaluation}

\begin{table*}[ht]
\caption{Number of required parameters in millions to train each model for the online graph snapshots/time steps}
\vspace{-0.2cm}
\centering
\begin{tabular}{c|c|c|c|c|c|c|c|c}

     & \multicolumn{8}{c}{\textbf{Yelp}} \\ \hline
     \textbf{Time Step} & \textbf{Distill2Vec}-$\mathcal{T}$ & \textbf{Distill2Vec}-$\mathcal{S}$ & \textbf{DynVGAE} & \textbf{DynamicTriad} & \textbf{TDGNN} & \textbf{DyREP} & \textbf{DMTKG}-$\mathcal{T}$ & \textbf{DMTKG}-$\mathcal{S}$ \\ \hline
     $1$ & $1.054$ & $\mathbf{0.214}$ & $6.090$   & $4.185$  & $2.593$ & $8.295$  & $2.182$ & $1.063$ \\\hline
     $2$ & $1.054$ & $\mathbf{0.238}$ & $6.649$   & $4.338$  & $2.984$ & $9.235$  & $2.182$ & $1.099$ \\\hline
     $3$ & $1.054$ & $\mathbf{0.261}$ & $7.187$   & $5.027$  & $3.495$ & $10.591$ & $2.182$ & $1.123$ \\\hline
     $4$ & $1.054$ & $\mathbf{0.283}$ & $7.685$   & $5.892$  & $3.891$ & $11.058$ & $2.182$ & $1.155$ \\\hline
     $5$ & $1.054$ & $\mathbf{0.304}$ & $8.225$  & $6.236$  & $4.185$ & $11.837$ & $2.182$ & $1.192$ \\\hline
     $6$ & $1.054$ & $\mathbf{0.327}$ & $8.809$  & $6.915$  & $4.563$ & $12.293$ & $2.182$ & $1.226$ \\\hline
     $7$ & $1.054$ & $\mathbf{0.351}$ & $9.380$  & $7.448$  & $4.982$ & $12.944$ & $2.182$ & $1.468$ \\\hline
     $8$ & $1.054$ & $\mathbf{0.375}$ & $9.933$  & $8.109$  & $5.527$ & $13.284$ & $2.182$ & $1.591$ \\\hline
     $9$ & $1.054$ & $\mathbf{0.398}$ & $10.308$ & $9.235$  & $6.019$ & $13.749$ & $2.182$ & $1.802$ \\\hline
     $10$ & $1.054$ & $\mathbf{0.413}$ & $10.658$ & $9.763$  & $6.237$ & $13.987$ & $2.182$ & $1.914$ \\\hline
     $11$ & $1.054$ & $\mathbf{0.428}$ & $11.236$ & $10.291$ & $6.832$ & $14.235$ & $2.182$ & $2.022$ \\ \hline \hline
     & \multicolumn{8}{c}{\textbf{ML-10M}} \\ \hline
     $1$ & $6.956$  & $\mathbf{1.542}$ & $6.035$  & $5.923$ & $4.285$ & $10.234$ & $5.293$ & $3.927$ \\\hline
     $2$ & $6.956$  & $\mathbf{1.700}$ & $6.668$  & $6.142$ & $4.928$ & $11.083$ & $5.293$ & $4.023$ \\\hline
     $3$ & $6.956$  & $\mathbf{2.011}$ & $7.911$  & $6.591$ & $5.291$ & $12.953$ & $5.293$ & $4.125$ \\\hline
     $4$ & $6.956$  & $\mathbf{2.127}$ & $8.374$  & $7.839$ & $6.018$ & $13.392$ & $5.293$ & $4.329$ \\\hline
     $5$ & $6.956$ & $\mathbf{2.264}$ & $8.922$  & $8.113$ & $6.827$ & $14.952$ & $5.293$ & $4.532$ \\\hline
     $6$ & $6.956$ & $\mathbf{2.375}$ & $9.367$  & $8.788$ & $7.283$ & $15.295$ & $5.293$ & $4.728$ \\\hline
     $7$ & $6.956$ & $\mathbf{2.562}$ & $10.113$ & $9.423$ & $8.183$ & $16.223$ & $5.293$ & $4.892$ \\
     
\end{tabular}

\label{tab:model_size}
\end{table*}

In Table \ref{tab:model_size}, we report the number of required  parameters in millions to train each model over the different online graph snapshots/time steps. As aforementioned in Section \ref{sec:model}, the teacher models Distill2Vec-$\mathcal{T}$ and DMTKG-$\mathcal{T}$ are trained on the offline data $\mathcal{G}^{\mathcal{T}}$. Therefore, the model sizes of Distill2Vec-$\mathcal{T}$ and DMTKG-$\mathcal{T}$ are not affected during the evaluation of the model on the online data $\mathcal{G}^{\mathcal{S}}$. We observe that Distill2Vec-$\mathcal{S}$ reduces the model size significantly, when compared with the teacher model Distill2Vec-$\mathcal{T}$, achieving averaged compression ratios of $31$:$100$ and $30$:$100$ for the Yelp and ML-10M datasets, respectively. Moreover, Distill2Vec-$\mathcal{S}$ constantly outperforms the baseline approaches in both datasets, in terms of the number of trainable parameters. We omit the number of parameters for Distill2Vec-$\mathcal{L}$, as it is a variant of Distill2Vec-$\mathcal{S}$ with equal number of parameters. The averaged compression ratios of Distill2Vec-$\mathcal{S}$ are $13$:$100$, $16$:$100$, $21$:$100$, $9$:$100$, $27$:$100$ and $37$:$100$, when evaluated against DynVGAE, DynamicTriad, TDGNN, DyREP, DMTKG-$\mathcal{T}$ and DMTKG-$\mathcal{S}$, respectively. The high compression ratios demonstrate the ability of our proposed distillation strategy to significantly reduce the number of model parameters. This means that the proposed student model Distill2Vec-$\mathcal{S}$ achieves low latency during the online inference of the node embeddings, compared with the other baseline approaches. We also notice that DyREP requires a large amount of trainable parameter in both datasets. This indicates that DyREP scales poorly when increasing the number of nodes in the graph, degrading the performance of the model for online graph snapshots. 

\begin{table*}[h]
    \caption{Average AUC for each online graph snapshots/time step}
    \vspace{-0.2cm}
    \centering
    \resizebox{\textwidth}{!}{
    \begin{tabular}{c|ccccccccc}
         & \multicolumn{9}{c}{\textbf{Yelp}} \\ \hline
         \textbf{Time Step} & \textbf{Distill2Vec}-$\mathcal{T}$ & \textbf{Distill2Vec}-$\mathcal{S}$ & \textbf{Distill2Vec}-$\mathcal{L}$ & \textbf{DynVGAE} & \textbf{DynamicTriad} &
         \textbf{TDGNN} &
         \textbf{DyREP} &
         \textbf{DMTKG}-$\mathcal{T}$ &
         \textbf{DMTKG}-$\mathcal{S}$ \\\hline 
         $1$  & $69.12 \pm 0.13$ & $\mathbf{69.23 \pm 0.12}$ & $69.13 \pm 0.12$ &$62.15 \pm 0.21$ & $67.32 \pm 0.10$ & $68.14 \pm 0.28$ & $64.17 \pm 0.05$ & $58.03 \pm 0.26$ & $59.42 \pm 0.29$\\\hline
         $2$  & $69.01 \pm 0.13$ & $\mathbf{69.32 \pm 0.11}$ & $69.15 \pm 0.14$ &$62.19 \pm 0.23$ & $67.41 \pm 0.09$ & $68.23 \pm 0.24$ & $64.86 \pm 0.04$ & $57.76 \pm 0.28$ & $58.72 \pm 0.23$\\\hline
         $3$  & $68.23 \pm 0.14$ & $\mathbf{69.38 \pm 0.11}$ & $69.19 \pm 0.12$ &$62.21 \pm 0.19$ & $67.12 \pm 0.09$ & $67.53 \pm 0.29$ & $65.58 \pm 0.02$ & $57.61 \pm 0.21$ & $58.94 \pm 0.26$\\\hline
         $4$  & $67.64 \pm 0.16$ & $\mathbf{69.68 \pm 0.14}$ & $69.21 \pm 0.12$ &$62.23 \pm 0.25$ & $67.58 \pm 0.06$ & $67.64 \pm 0.25$ & $65.82 \pm 0.06$ & $57.44 \pm 0.27$ & $57.83 \pm 0.27$\\\hline
         $5$ & $66.97 \pm 0.15$ & $\mathbf{69.89 \pm 0.11}$ & $69.25 \pm 0.13$ &$62.22 \pm 0.25$ & $67.93 \pm 0.08$ & $68.18 \pm 0.25$ & $65.91 \pm 0.02$ & $54.89 \pm 0.28$ & $56.28 \pm 0.27$\\\hline
         $6$ & $65.59 \pm 0.14$ & $\mathbf{69.92 \pm 0.11}$ & $69.27 \pm 0.12$ &$62.25 \pm 0.24$ & $67.24 \pm 0.11$ & $69.19 \pm 0.26$ & $66.32 \pm 0.05$ & $55.27 \pm 0.29$ & $56.63 \pm 0.28$\\\hline
         $7$ & $65.02 \pm 0.16$ & $\mathbf{70.01 \pm 0.10}$ & $69.32 \pm 0.13$ &$62.35 \pm 0.23$ & $68.62 \pm 0.09$ & $68.76 \pm 0.23$ & $66.57 \pm 0.04$ & $55.11 \pm 0.25$ & $57.69 \pm 0.24$\\\hline
         $8$ & $64.54 \pm 0.17$ & $\mathbf{70.01 \pm 0.11}$ & $69.36 \pm 0.12$ &$62.46 \pm 0.24$ & $68.82 \pm 0.08$ & $69.09 \pm 0.27$ & $66.54 \pm 0.06$ & $55.03 \pm 0.26$ & $56.14 \pm 0.28$\\\hline
         $9$ & $64.09 \pm 0.15$ & $\mathbf{70.03 \pm 0.11}$ & $69.41 \pm 0.14$ &$62.82 \pm 0.25$ & $68.89 \pm 0.10$ & $69.06 \pm 0.28$ & $67.78 \pm 0.06$ & $56.40 \pm 0.28$ & $58.68 \pm 0.24$\\\hline
         $10$ & $64.01 \pm 0.16$ & $\mathbf{69.96 \pm 0.12}$ & $69.52 \pm 0.13$ &$62.91 \pm 0.21$ & $68.92 \pm 0.09$ & $68.58 \pm 0.24$ & $67.51 \pm 0.07$ & $55.32 \pm 0.26$ & $59.49 \pm 0.26$\\\hline
         $11$ & $63.25 \pm 0.17$ & $\mathbf{69.12 \pm 0.13}$ & $68.84 \pm 0.12$ &$63.02 \pm 0.22$ & $68.15 \pm 0.07$ & $68.26 \pm 0.24$ & $66.40 \pm 0.09$ & $54.95 \pm 0.29$ & $60.19 \pm 0.28$\\ \hline \hline 
     & \multicolumn{9}{c}{\textbf{ML-10M}} \\ \hline
         $1$  & $90.94 \pm 0.34$ & $\mathbf{90.95 \pm 0.26}$ & $90.95 \pm 0.12$ &$79.43 \pm 0.52$ & $86.63 \pm 0.41$ & $88.35 \pm 0.52$ & $83.49 \pm 0.41$ & $72.42 \pm 0.19$ & $73.19 \pm 0.17$ \\\hline
         $2$  & $90.42 \pm 0.39$ & $\mathbf{91.53 \pm 0.22}$ & $90.97 \pm 0.13$ &$80.15 \pm 0.61$ & $87.86 \pm 0.42$ & $88.89 \pm 0.51$ & $83.92 \pm 0.44$ & $73.64 \pm 0.12$ & $75.82 \pm 0.16$ \\\hline
         $3$  & $89.92 \pm 0.31$ & $\mathbf{92.68 \pm 0.25}$ & $91.04 \pm 0.13$ &$80.37 \pm 0.56$ & $87.91 \pm 0.45$ & $89.26 \pm 0.52$ & $85.02 \pm 0.45$ & $73.82 \pm 0.14$ & $75.53 \pm 0.18$ \\\hline
         $4$  & $89.84 \pm 0.30$ & $\mathbf{93.26 \pm 0.27}$ & $91.13 \pm 0.11$ &$81.02 \pm 0.58$ & $88.23 \pm 0.45$ & $90.64 \pm 0.55$ & $86.25 \pm 0.44$ & $74.03 \pm 0.15$ & $75.67 \pm 0.14$ \\\hline
         $5$ & $88.69 \pm 0.38$ & $\mathbf{94.14 \pm 0.25}$ & $92.37 \pm 0.11$ &$82.64 \pm 0.51$ & $89.56 \pm 0.44$ & $92.20 \pm 0.52$ & $85.98 \pm 0.46$ & $73.76 \pm 0.14$ & $75.82 \pm 0.15$ \\\hline
         $6$ & $88.29 \pm 0.32$ & $\mathbf{94.47 \pm 0.21}$ & $92.59 \pm 0.12$ &$82.86 \pm 0.45$ & $90.86 \pm 0.45$ & $92.45 \pm 0.53$ & $86.14 \pm 0.42$ & $74.21 \pm 0.16$ & $75.74 \pm 0.16$ \\\hline
         $7$ & $87.58 \pm 0.37$ & $\mathbf{94.69 \pm 0.28}$ & $92.84 \pm 0.12$ &$82.91 \pm 0.59$ & $90.94 \pm 0.42$ & $92.61 \pm 0.52$ & $87.01 \pm 0.45$ & $73.97 \pm 0.17$ & $76.18 \pm 0.17$ \\
    \end{tabular}
}
    \label{tab:auc}
\end{table*}

\begin{table*}[ht]
\caption{Impact on AUC when varying the size of the embedding dimension $d$ in the proposed Distill2Vec models. The reported values are averaged over all the online time steps.}
\vspace{-0.2cm}
    \centering
    \begin{tabular}{|c|c|cccccc|}
    \hline
         \textbf{Dataset} & \textbf{Model} & \multicolumn{6}{c|}{\textbf{Embedding Size $d$}} \\ \hline \hline
         & & $16$ & $32$ & $64$ & $128$ & $256$ & $512$ \\\hline
         \multirow{2}{*}{\textbf{Yelp}} & \textbf{Distill2Vec}-$\mathcal{T}$ & $62.12 \pm 0.14$ & $63.72 \pm 0.13$ & $64.44 \pm 0.13$ & $65.98 \pm 0.14$ & $\mathbf{66.13 \pm 0.12}$ & $65.01 \pm 0.13$ \\
         & \textbf{Distill2Vec}-$\mathcal{S}$ & $67.92 \pm 0.13$ & $\mathbf{69.69 \pm 0.12}$ & $68.89 \pm 0.13$ & $68.26 \pm 0.12$ & $68.22 \pm 0.13$ & $68.13 \pm 0.14$ \\ \hline
         \multirow{2}{*}{\textbf{ML-10M}} & \textbf{Distill2Vec}-$\mathcal{T}$ & $85.43 \pm 0.36$ & $86.16 \pm 0.33$ & $87.89 \pm 0.33$ & $89.01 \pm 0.31$ & $\mathbf{89.38 \pm 0.34}$ &  $89.32 \pm 0.32$\\
         & \textbf{Distill2Vec}-$\mathcal{S}$ & $87.04 \pm 0.24$ & $88.32 \pm 0.26$ & $\mathbf{93.10 \pm 0.26}$ & $90.01 \pm 0.24$ &  $89.85 \pm 0.23$ & $89.02 \pm 0.24$ \\ \hline
    \end{tabular}
    
    \label{tab:embeddings}
\end{table*}

\begin{table*}[ht]
    \caption{Impact on AUC when varying the number of attention heads $h$ and $g$ in the proposed Distill2Vec models.}
    \vspace{-0.2cm}
    \centering
    \begin{tabular}{|c|c|ccccc|}
    \hline
         \textbf{Dataset} & \textbf{Model} & \multicolumn{5}{c|}{\textbf{Attention Heads $h$/$g$}} \\ \hline \hline
         & & $2$ & $4$ & $8$ & $16$ & $32$ \\\hline
         \multirow{2}{*}{\textbf{Yelp}} & \textbf{Distill2Vec}-$\mathcal{T}$ & $64.92 \pm 0.12$ & $65.24 \pm 0.13$ & $65.59 \pm 0.13$ & $\mathbf{66.13 \pm 0.12}$ & $65.88 \pm 0.12$ \\
         & \textbf{Distill2Vec}-$\mathcal{S}$ & $\mathbf{69.69 \pm 0.12}$ & $67.94 \pm 0.13$ & $67.49 \pm 0.12$ & $67.91 \pm 0.13$ & $67.08 \pm 0.12$ \\ \hline
         \multirow{2}{*}{\textbf{ML-10M}} & \textbf{Distill2Vec}-$\mathcal{T}$ & $87.82 \pm 0.32$ & $88.76 \pm 0.34$ & $\mathbf{89.38 \pm 0.34}$ & $88.83 \pm 0.34$ & $88.24 \pm 0.32$ \\
         & \textbf{Distill2Vec}-$\mathcal{S}$ & $88.23 \pm 0.26$ & $\mathbf{93.10 \pm 0.26}$ & $89.92 \pm 0.24$ & $89.23 \pm 0.25$ & $89.11 \pm 0.23$ \\ \hline
    \end{tabular}
    \label{tab:attentions}
\end{table*}

In Table \ref{tab:auc}, we evaluate the performance of the student model Distill2Vec-$\mathcal{S}$ against the baseline approaches in the link prediction task. We observe that the student model Distill2Vec-$\mathcal{S}$ constantly outperforms the baseline approaches, in terms of AUC, for both datasets. This indicates that the proposed knowledge distillation strategy can efficiently transfer the knowledge of the pretrained model Distill2Vec-$\mathcal{T}$ to the student model Distill2Vec-$\mathcal{S}$. Therefore, Distill2Vec-$\mathcal{S}$ achieves high link prediction accuracy, while reducing the number of trainable parameters. Moreover, we observe that Distill2Vec-$\mathcal{L}$ exhibits similar behaviour as Distill2Vec-$\mathcal{S}$. However, the cross entropy function in Distill2Vec-$\mathcal{L}$ limits the prediction accuracy, when compared with the Kullback-Leibler divergence of the proposed the Distill2Vec-$\mathcal{S}$ model. Evaluated against TDGNN, which is the second best baseline approach in all datasets, Distill2Vec-$\mathcal{S}$ achieves relative gains $1.8$ and $2.5\%$ for the Yelp and ML-10M datasets, respectively. Note that as shown in Table~\ref{tab:model_size} Distill2Vec-$\mathcal{S}$ achieves average compression ratio of $7$:$100$ and $35$:$100$, in terms of trainable parameters, when compared with TDGNN for the Yelp and ML-10M dataset, respectively. Thus, our model is able to capture the evolution of the graph in the learned node representations, while significantly reducing the model size.
 
In addition, on inspection of Table \ref{tab:auc} we observe that the student models Distill2Vec-$\mathcal{S}$ and DMTKG-$\mathcal{S}$ constantly outperform their respective teacher models Distill2Vec-$\mathcal{T}$ and DMTKG-$\mathcal{T}$. This demonstrates the capability of student models to overcome any bias introduced by the pretrained teacher models on the offline data. Thus, the student model Distill2Vec-$\mathcal{S}$ achieves relative gains of $5.5$ and $4.2\%$ against the teacher model Distill2Vec-$\mathcal{T}$ for the Yelp and ML-10M datasets, respectively.

\subsection{Parameter Sensitivity} \label{sec:param_sens}

In Table \ref{tab:embeddings}, we compare the proposed Distill2Vec-$\mathcal{T}$ and Distill2Vec-$\mathcal{S}$ models in terms of AUC when varying the node embedding sizes $d$ in the range of $\{16,32,64,128,256,512\}$. We observe that the teacher model Distill2Vec-$\mathcal{T}$ requires higher embedding sizes than the student model Distill2Vec-$\mathcal{S}$. This demonstrates the effectiveness of the knowledge distillation strategy to transfer the knowledge of the pretrained teacher model Distill2Vec-$\mathcal{T}$ to the student model Distill2Vec-$\mathcal{S}$. Therefore, the student model Distill2Vec-$\mathcal{S}$ mimics the accurate node embeddings produced by the teacher model, by requiring less fine-grained representations to achieve high prediction accuracy \cite{Jiaxi2018, hinton2015, Bucilua2006}. 

In Table \ref{tab:attentions}, we evaluate the influence of the number of structural attention heads $h$ and temporal attention heads $g$ on the link prediction accuracy of the Distill2Vec-$\mathcal{T}$ and Distill2Vec-$\mathcal{S}$ models. For fair comparison, we fix equal number of attention heads in both the structural and attention layers. High values of attention heads allow the Distill2Vec-$\mathcal{T}$ and Distill2Vec-$\mathcal{S}$ models to capture different latent facets for each node in the graphs (Section \ref{sec:teacher}). In both datasets, Distill2Vec-$\mathcal{S}$ requires less number of attention heads than the teacher model Distill2Vec-$\mathcal{T}$. Considering that the teacher model is trained on the offline data, Distill2Vec-$\mathcal{T}$ achieves high prediction accuracy with high number of attention heads. Thereafter, the student model Distill2Vec-$\mathcal{S}$ distills the different latent facets from the teacher model Distill2Vec-$\mathcal{T}$ and learns accurate node embeddings with low number of attention heads.

\begin{figure}[h] \centering
\begin{tabular}{cc}
\includegraphics[scale=0.112]{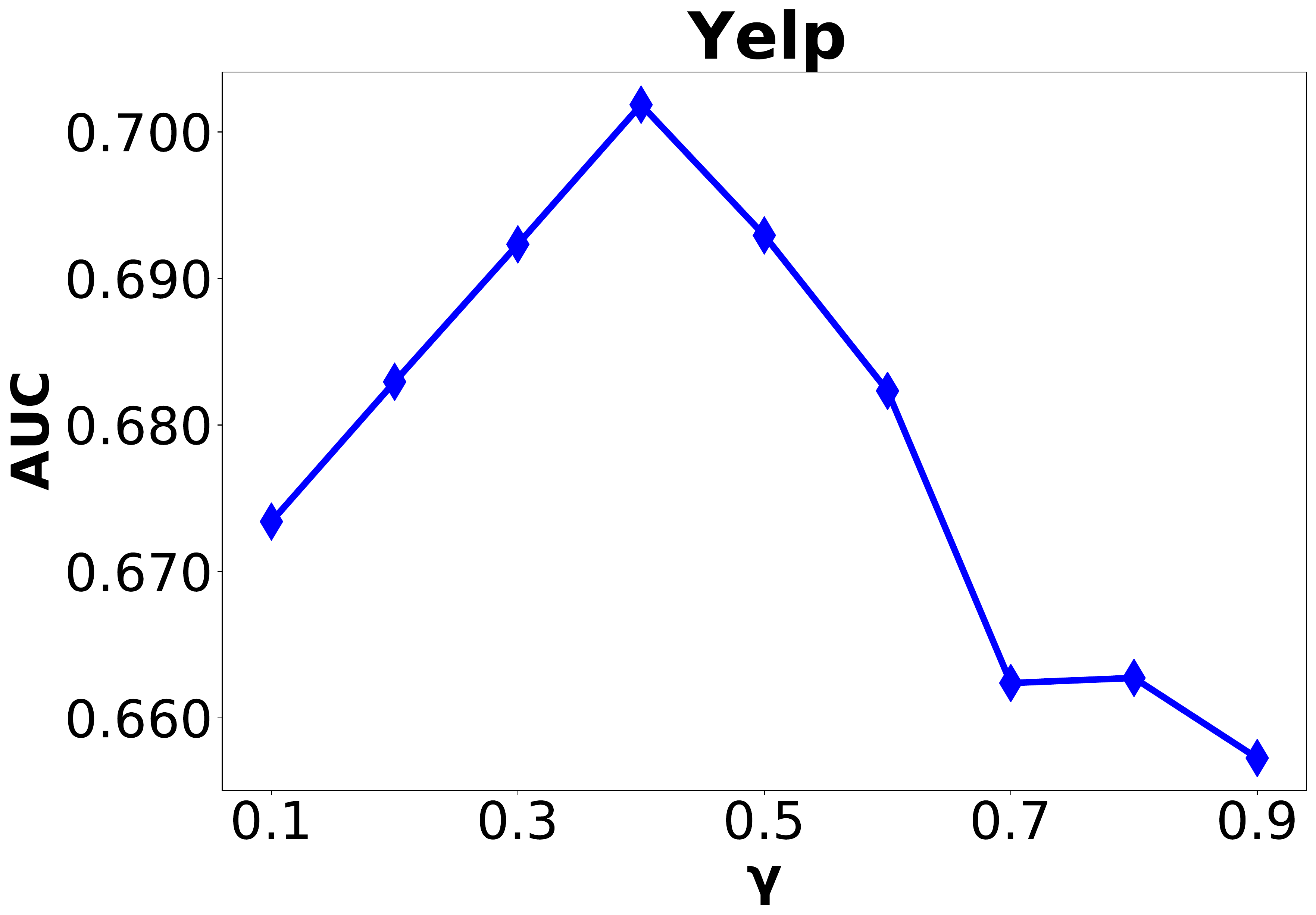} &
\includegraphics[scale=0.112]{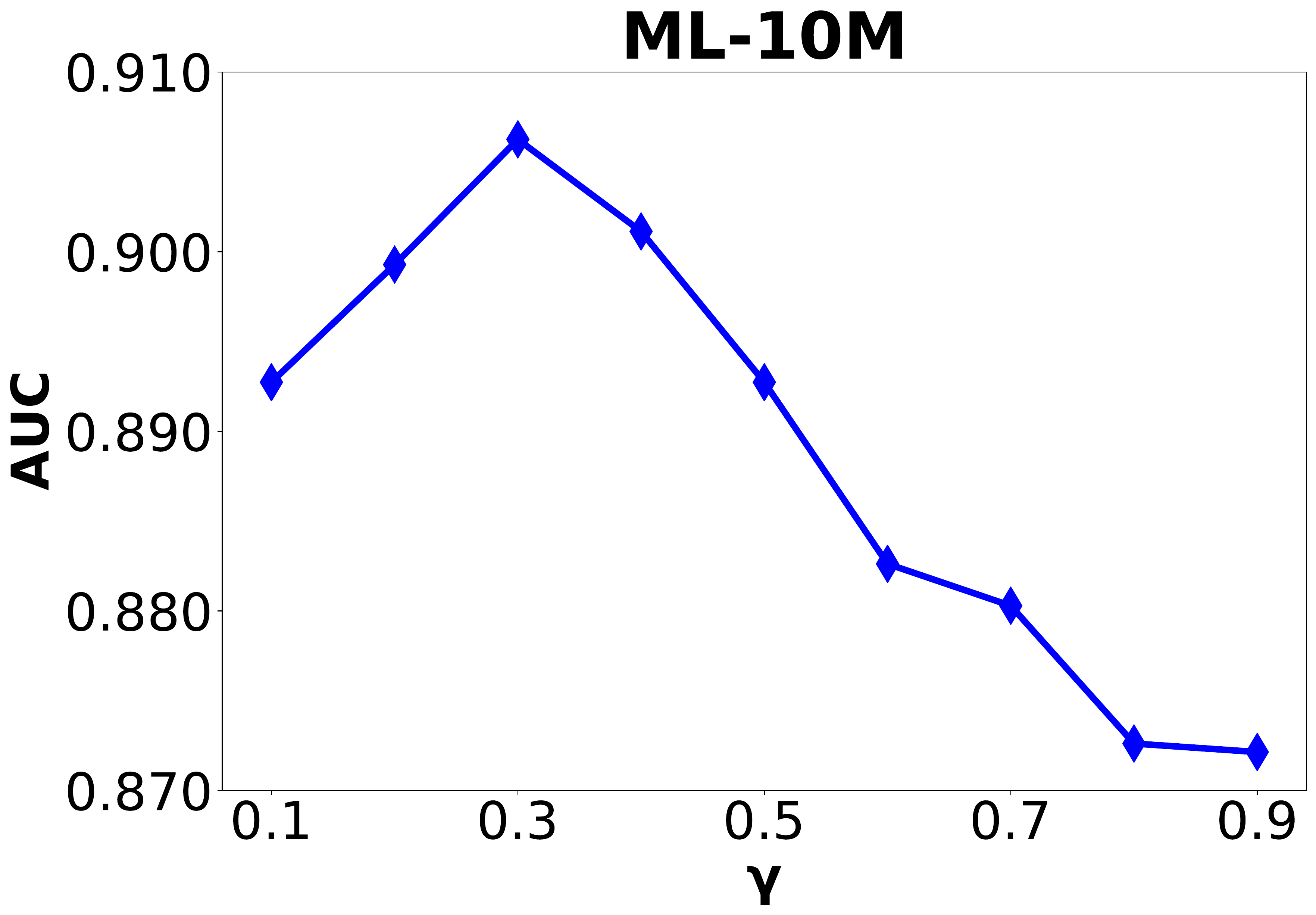}
\end{tabular}
\vspace{-0.2cm}
\caption{Impact of $\gamma$ on the prediction accuracy of Distill2Vec-$\mathcal{S}$} \label{fig:student_gamma}
\end{figure}

In Figure \ref{fig:student_gamma}, we study the impact of the hyperparameter $\gamma$ (Equation \ref{eq:distil}) on the performance of the student model Distill2Vec-$\mathcal{S}$. For each value of $\gamma$ we report average AUC for Distill2Vec-$\mathcal{S}$ on the online data $\mathcal{G}^{\mathcal{S}}$ over all the time steps. The best $\gamma$ values are 0.4 and 0.3 in Yelp and ML-10M, respectively. In both datasets, the performance grows linearly for $\gamma\leq 0.3$. Instead, high values of $\gamma$ degrade the performance of Distill2Vec-$\mathcal{S}$, as the student model Distill2Vec-$\mathcal{S}$ emphasizes more on the loss $L^\mathcal{S}$ in Equation \ref{eq:distil} and distills less knowledge by the teacher model Distill2Vec-$\mathcal{T}$. This occurs because for high values of $\gamma$ the student model Distill2Vec-$\mathcal{S}$ is trained based on the prediction error of $L^{\mathcal{T}}$ than the loss $L^{\mathcal{S}}$. This means that the training of the student model Distill2Vec-$\mathcal{S}$  is mainly supervised by the teacher model, discarding any further training on the online data. Instead, decreasing the hyperparameter $\gamma$ prevents the student model Distill2Vec-$\mathcal{S}$ to distill the knowledge of the teacher model. The student model Distill2Vec-$\mathcal{S}$ learns node embeddings based on the prediction accuracy of the model on the online data, disregarding the knowledge of the teacher model Distill2Vec-$\mathcal{T}$. 

\begin{figure}[h] \centering
\begin{tabular}{cc}
\includegraphics[scale=0.112]{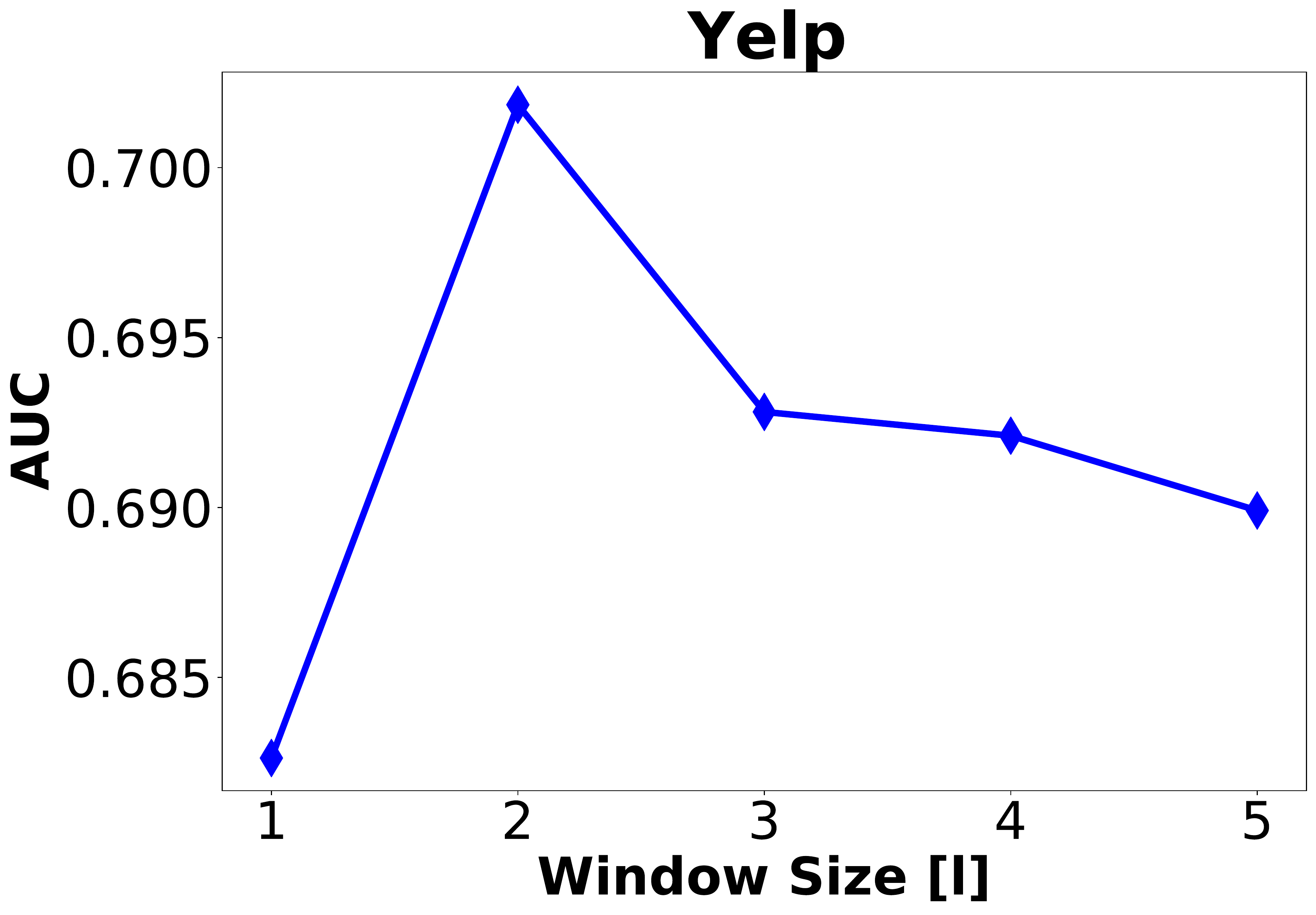} &
\includegraphics[scale=0.112]{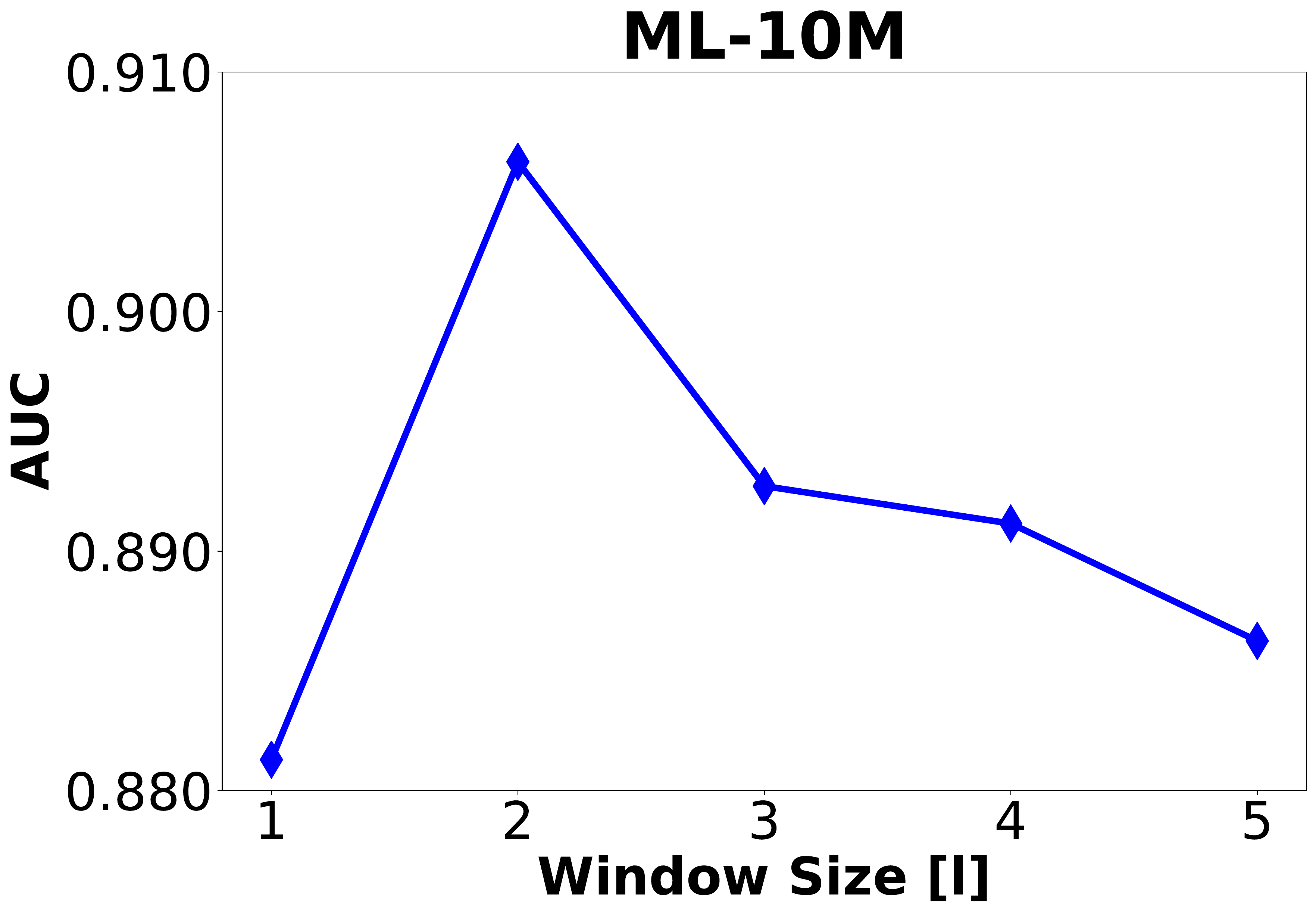}
\end{tabular}

\vspace{-0.2cm}
\caption{Impact of window size $l$ on the prediction accuracy of Distill2Vec-$\mathcal{S}$} \label{fig:student_window}
\end{figure}

In Figure \ref{fig:student_window}, we present the impact of the window size $l$ on the link prediction performance of the student model Distill2Vec-$\mathcal{S}$. We vary the window size $l$ from $1$ to $5$ by a step of 1. We report the average AUC of the student model over all the graph snapshots of the online data. Distill2Vec-$\mathcal{S}$ achieves the highest performance when setting $l=2$ previous graph snapshots. Increasing the window size to $l > 2$ negatively impacts the performance of the Distill2Vec-$\mathcal{S}$, as more graph snapshots introduce noise during the training of the model. This observation reflects on the highly evolving nature of the graphs, where bursty events such as new movie release, restaurant opening, and so on, cause significant differences between consecutive graph snapshots \cite{Sankar2020}.

\section{Conclusion} \label{sec:conclusion}

In this paper, we presented a knowledge distillation strategy to reduce the size of a teacher model for dynamic graph representation learning. The proposed distillation strategy can efficiently generate a compact student model with low online inference latency, while achieving high link prediction accuracy. The experimental results demonstrate the compression efficiency of our distillation strategy. The proposed student model achieves a compression ratio up to $31$:$100$ on two real-world datasets, when compared with the pretrained teacher model. Evaluated against several state-of-the-art approaches, the proposed student model achieves an average relative improvement of $2.2\%$ on both datasets, by significantly reducing the number of required parameters. An interesting future direction is to explore the performance of data-free distillation strategies on dynamic graph representation learning approaches \cite{Hanting2019, MicaelliS19}. The main challenge is to design the student model so as to infer accurate embeddings on unobserved nodes by the teacher model.

\bibliographystyle{IEEEtran}
\bibliography{IEEEexample}

\begin{thebibliography}{10}
\providecommand{\url}[1]{#1}
\csname url@samestyle\endcsname
\providecommand{\newblock}{\relax}
\providecommand{\bibinfo}[2]{#2}
\providecommand{\BIBentrySTDinterwordspacing}{\spaceskip=0pt\relax}
\providecommand{\BIBentryALTinterwordstretchfactor}{4}
\providecommand{\BIBentryALTinterwordspacing}{\spaceskip=\fontdimen2\font plus
\BIBentryALTinterwordstretchfactor\fontdimen3\font minus
  \fontdimen4\font\relax}
\providecommand{\BIBforeignlanguage}[2]{{%
\expandafter\ifx\csname l@#1\endcsname\relax
\typeout{** WARNING: IEEEtran.bst: No hyphenation pattern has been}%
\typeout{** loaded for the language `#1'. Using the pattern for}%
\typeout{** the default language instead.}%
\else
\language=\csname l@#1\endcsname
\fi
#2}}
\providecommand{\BIBdecl}{\relax}
\BIBdecl

\bibitem{Zhu2017}
L.~Zhu, D.~Guo, J.~Yin, G.~V. Steeg, and A.~Galstyan, ``Scalable temporal
  latent space inference for link prediction in dynamic social networks
  (extended abstract),'' in \emph{ICDE}, 2017, pp. 57--58.

\bibitem{velickovic2018}
P.~Velickovic, G.~Cucurull, A.~Casanova, A.~Romero, P.~Li{\`{o}}, and
  Y.~Bengio, ``Graph attention networks,'' in \emph{ICLR}, 2018.

\bibitem{hamilton2017}
W.~L. Hamilton, Z.~Ying, and J.~Leskovec, ``Inductive representation learning
  on large graphs,'' in \emph{NIPS}, 2017, pp. 1024--1034.

\bibitem{Fout2017}
A.~Fout, J.~Byrd, B.~Shariat, and A.~Ben-Hur, ``Protein interface prediction
  using graph convolutional networks,'' in \emph{NIPS}, 2017, p. 6533–6542.

\bibitem{Cao2019}
Y.~Cao, X.~Wang, X.~He, Z.~Hu, and T.-S. Chua, ``Unifying knowledge graph
  learning and recommendation: Towards a better understanding of user
  preferences,'' in \emph{WWW}, 2019, p. 151–161.

\bibitem{Goyal2018ht}
P.~Goyal, A.~Sapienza, and E.~Ferrara, ``Recommending teammates with deep
  neural networks,'' in \emph{HT}, 2018, p. 57–61.

\bibitem{hamilton17}
W.~L. Hamilton, R.~Ying, and J.~Leskovec, ``Representation learning on graphs:
  Methods and applications,'' \emph{{IEEE} Data Eng. Bull.}, vol.~40, no.~3,
  pp. 52--74, 2017.

\bibitem{Liu2018}
N.~Liu, X.~Huang, J.~Li, and X.~Hu, ``On interpretation of network embedding
  via taxonomy induction,'' in \emph{KDD}, 2018, p. 1812–1820.

\bibitem{Geng2015}
X.~{Geng}, H.~{Zhang}, J.~{Bian}, and T.~{Chua}, ``Learning image and user
  features for recommendation in social networks,'' in \emph{ICCV}, 2015, pp.
  4274--4282.

\bibitem{Zhao2019}
Y.~Zhao, X.~Wang, H.~Yang, L.~Song, and J.~Tang, ``Large scale evolving graphs
  with burst detection,'' in \emph{IJCAI}, 2019, pp. 4412--4418.

\bibitem{Grover2016}
A.~Grover and J.~Leskovec, ``node2vec: Scalable feature learning for
  networks,'' in \emph{KDD}, 2016, pp. 855--864.

\bibitem{kipf2017}
T.~N. Kipf and M.~Welling, ``Semi-supervised classification with graph
  convolutional networks,'' in \emph{ICLR}, 2017.

\bibitem{Perozzi2014}
B.~Perozzi, R.~Al-Rfou, and S.~Skiena, ``Deepwalk: Online learning of social
  representations,'' in \emph{KDD}, 2014, pp. 701--710.

\bibitem{Sankar2020}
A.~Sankar, Y.~Wu, L.~Gou, W.~Zhang, and H.~Yang, ``Dysat: Deep neural
  representation learning on dynamic graphs via self-attention networks,'' in
  \emph{WSDM}, 2020, pp. 519--527.

\bibitem{Zhou2018}
L.~Zhou, Y.~Yang, X.~Ren, F.~Wu, and Y.~Zhuang, ``Dynamic network embedding by
  modeling triadic closure process,'' in \emph{AAAI}, 2018, pp. 571--578.

\bibitem{Trivedi2019DyRepLR}
R.~Trivedi, M.~Farajtabar, P.~Biswal, and H.~Zha, ``Dyrep: Learning
  representations over dynamic graphs,'' in \emph{ICLR}, 2019.

\bibitem{mahdavi2019}
S.~Mahdavi, S.~Khoshraftar, and A.~An, ``Dynamic joint variational graph
  autoencoders,'' in \emph{ECML}, 2019, pp. 385--401.

\bibitem{Hajiramezanali2019}
E.~Hajiramezanali, A.~Hasanzadeh, K.~R. Narayanan, N.~Duffield, M.~Zhou, and
  X.~Qian, ``Variational graph recurrent neural networks,'' in \emph{NeurIPS},
  2019, pp. 10\,700--10\,710.

\bibitem{Bucilua2006}
C.~Bucila, R.~Caruana, and A.~Niculescu{-}Mizil, ``Model compression,'' in
  \emph{KDD}, 2006, pp. 535--541.

\bibitem{hinton2015}
G.~Hinton, O.~Vinyals, and J.~Dean, ``Distilling the knowledge in a neural
  network,'' in \emph{NIPS}, 2015.

\bibitem{Liu_2019_CVPR}
Y.~Liu, J.~Cao, B.~Li, C.~Yuan, W.~Hu, Y.~Li, and Y.~Duan, ``Knowledge
  distillation via instance relationship graph,'' in \emph{CVPR}, 2019, pp.
  7096--7104.

\bibitem{Mary2019}
M.~Phuong and C.~Lampert, ``Towards understanding knowledge distillation,'' in
  \emph{ICML}, 2019, pp. 5142--5151.

\bibitem{Jiaxi2018}
J.~Tang and K.~Wang, ``Ranking distillation: Learning compact ranking models
  with high performance for recommender system,'' in \emph{KDD}, 2018, p.
  2289–2298.

\bibitem{Li2017}
H.~Li, T.~N. Chan, M.~L. Yiu, and N.~Mamoulis, ``Fexipro: Fast and exact inner
  product retrieval in recommender systems,'' in \emph{SIGMOD}, 2017, p.
  835–850.

\bibitem{Cao2015}
S.~Cao, W.~Lu, and Q.~Xu, ``Grarep: Learning graph representations with global
  structural information,'' in \emph{CIKM}, 2015, p. 891–900.

\bibitem{Ou2016}
M.~Ou, P.~Cui, J.~Pei, Z.~Zhang, and W.~Zhu, ``Asymmetric transitivity
  preserving graph embedding,'' in \emph{KDD}, 2016, p. 1105–1114.

\bibitem{kipf2016variational}
A.~Hasanzadeh, E.~Hajiramezanali, K.~R. Narayanan, N.~Duffield, M.~Zhou, and
  X.~Qian, ``Semi-implicit graph variational auto-encoders,'' in
  \emph{NeurIPS}, 2019, pp. 10\,711--10\,722.

\bibitem{Sarkar2005}
P.~Sarkar and A.~W. Moore, ``Dynamic social network analysis using latent space
  models,'' \emph{SIGKDD}, vol.~7, no.~2, 2005.

\bibitem{goyal2018dyngem}
P.~Goyal, N.~Kamra, X.~He, and Y.~Liu, ``Dyngem: Deep embedding method for
  dynamic graphs,'' \emph{arXiv preprint arXiv:1805.11273}, 2018.

\bibitem{Pareja2020}
A.~Pareja, G.~Domeniconi, J.~Chen, T.~Ma, T.~Suzumura, H.~Kanezashi, T.~Kaler,
  T.~B. Schardl, and C.~E. Leiserson, ``{EvolveGCN}: Evolving graph
  convolutional networks for dynamic graphs,'' in \emph{AAAI}, 2020.

\bibitem{Goyal2020}
P.~Goyal, S.~R. Chhetri, and A.~Canedo, ``dyngraph2vec: Capturing network
  dynamics using dynamic graph representation learning,'' \emph{Knowl. Based
  Syst.}, vol. 187, 2020.

\bibitem{Jiaqi2019}
J.~Ma and Q.~Mei, ``Graph representation learning via multi-task knowledge
  distillation,'' in \emph{NeurIPS}, 2019.

\bibitem{Goyal2018}
P.~Goyal, N.~Kamra, X.~He, and Y.~Liu, ``Dyngem: Deep embedding method for
  dynamic graphs,'' vol. abs/1805.11273, 2018.

\bibitem{anil2018}
R.~Anil, G.~Pereyra, A.~Passos, R.~Orm{\'{a}}ndi, G.~E. Dahl, and G.~E. Hinton,
  ``Large scale distributed neural network training through online
  distillation,'' in \emph{ICLR}, 2018.

\bibitem{Jimmy2014}
J.~Ba and R.~Caruana, ``Do deep nets really need to be deep?'' in \emph{NIPS},
  2014, pp. 2654--2662.

\bibitem{vaswani2017attention}
A.~Vaswani, N.~Shazeer, N.~Parmar, J.~Uszkoreit, L.~Jones, A.~N. Gomez,
  {\L}.~Kaiser, and I.~Polosukhin, ``Attention is all you need,'' in
  \emph{Advances in neural information processing systems}, 2017, pp.
  5998--6008.

\bibitem{Gehring2017}
J.~Gehring, M.~Auli, D.~Grangier, D.~Yarats, and Y.~N. Dauphin, ``Convolutional
  sequence to sequence learning,'' in \emph{ICML}, 2017, pp. 1243--1252.

\bibitem{Tian2020Contrastive}
Y.~Tian, D.~Krishnan, and P.~Isola, ``Contrastive representation
  distillation,'' in \emph{ICLR}, 2020.

\bibitem{Qu2020}
L.~Qu, H.~Zhu, Q.~Duan, and Y.~Shi, ``Continuous-time link prediction via
  temporal dependent graph neural network,'' in \emph{WWW}, 2020, p.
  3026–3032.

\bibitem{li2016deepgraph}
C.~Li, X.~Guo, and Q.~Mei, ``Deepgraph: Graph structure predicts network
  growth,'' 2016.

\bibitem{egad2020}
S.~Antaris, D.~Rafailidis, and S.~Girdzijauskas, ``{EGAD}: Evolving graph
  representation learning with self-attention and knowledge distillation for
  live video streaming events,'' in \emph{IEEE Big Data}, 2020.

\bibitem{kingma2014}
D.~P. Kingma and J.~Ba, ``Adam: {A} method for stochastic optimization,'' in
  \emph{ICLR}, 2015.

\bibitem{Hanting2019}
H.~Chen, Y.~Wang, C.~Xu, Z.~Yang, C.~Liu, B.~Shi, C.~Xu, C.~Xu, and Q.~Tian,
  ``Data-free learning of student networks,'' in \emph{ICCV}, 2019, pp.
  3513--3521.

\bibitem{MicaelliS19}
P.~Micaelli and A.~J. Storkey, ``Zero-shot knowledge transfer via adversarial
  belief matching,'' in \emph{NeurIPS}, 2019, pp. 9547--9557.

\end{thebibliography}
\end{document}